\documentclass[final]{cvpr}
 \usepackage{times}
\usepackage{epsfig}

\usepackage{graphicx}
\usepackage{amsmath}
\usepackage{amssymb}
\usepackage{booktabs}
\usepackage{algorithm}
\usepackage{algorithmic}
\usepackage{multirow}
\usepackage{subcaption}
\usepackage{makecell}
\usepackage{adjustbox}
\usepackage{threeparttable}
\usepackage{bbding}
\usepackage[table,xcdraw]{xcolor}

\usepackage{pifont}
\usepackage[pagebackref=true,breaklinks=true,colorlinks,bookmarks=false]{hyperref}

\def\cvprPaperID{708} % *** Enter the CVPR Paper ID here
\def\confYear{CVPR 2021}

\begin{document}

%%%%%%%%% TITLE
\title{TSI: Temporal Saliency Integration for Video Action Recognition
\vspace{-0.4cm}
}
\author{Haisheng Su$^{1\dag}$\quad 
Kunchang Li$^{2}$\quad 
Jinyuan Feng$^{3}$ \quad
Dongliang Wang$^{1}$ \quad \\
Weihao Gan$^{1}$ \quad
Wei Wu$^{1}$ \quad 
Yu Qiao$^{2,4}$ \\
$^{1}${SenseTime Research} \\ $^{2}$Shenzhen Institutes of Advanced Technology, Chinese Academy of Sciences\\ $^{3}${Chongqing University} \\
$^{4}$Shanghai AI Laboratory, Shanghai, China\\
{\tt\small \{suhaisheng,wangdongliang,ganweihao,wuwei\}@sensetime.com}\\
{\tt\small  lkc4217@gmail.com, fengjinyuan@outlook.com, yu.qiao@siat.ac.cn}
\vspace{-0.4cm}
}

\maketitle
\pagestyle{empty}  % no page number for the second and the later pages
\thispagestyle{empty} % no page number for the first page
\let\thefootnote\relax\footnotetext{$\dag$ Corresponding author.}
%%%%%%%%% ABSTRACT

\begin{abstract}
Efficient spatiotemporal modeling is an important yet challenging problem for video action recognition. Existing state-of-the-art methods exploit neighboring feature differences to obtain motion clues for short-term temporal modeling with a simple convolution. However, only one local convolution is incapable of handling various kinds of actions because of the limited receptive field. Besides, action-irrelated noises brought by camera movement will also harm the quality of extracted motion features. In this paper, we propose a Temporal Saliency Integration (TSI) block, which mainly contains a Salient Motion Excitation (SME) module and a Cross-perception Temporal Integration (CTI) module. Specifically, SME aims to highlight the motion-sensitive area through spatial-level local-global motion modeling, where the saliency alignment and pyramidal motion modeling are conducted successively between adjacent frames to capture motion dynamics with fewer noises caused by misaligned background. CTI is designed to perform multi-perception temporal modeling through a group of separate 1D convolutions respectively. Meanwhile, temporal interactions across different perceptions are integrated with the attention mechanism. Through these two modules, long short-term temporal relationships can be encoded efficiently by introducing limited additional parameters. Extensive experiments are conducted on several popular benchmarks (i.e., Something-Something V1 \& V2, Kinetics-400, UCF-101, and HMDB-51), which demonstrate the effectiveness of our proposed method.
\end{abstract}

%Existing state-of-the-art methods exploit motion clues to assist in short-term temporal modeling through feature difference over consecutive frames.
%However, insignificant noises will be inevitably introduced due to the camera movement. 

%%%%%%%%% BODY TEXT
\section{Introduction}
\label{sec:intro}
%\textcolor{red}{simply introduce the history, mainly focus on 2D CNN and raise the key problem to be resolve: reduce action-irrelated information}

% With the rapid advancement of Internet and intelligent cameras, video data have increased dramatically over the past years. Hence, efficient video analytic technology is important and attracts many research interests in video understanding. Action recognition, which aims to classify the trimmed videos into specific categories, is one of the basic tasks in video analytics. And it has a wide range of applications such as intelligent surveillance, human-computer interaction, and personalized recommendation. Undoubtedly, efficient spatiotemporal modeling of videos is vital for achieving promising recognition accuracy and meanwhile ensuring reasonable inference complexity.

\begin{figure}[t]
\centering
\setlength{\abovecaptionskip}{-0.cm} %缩小caption和图像之间的距离
\includegraphics[width=1\columnwidth]{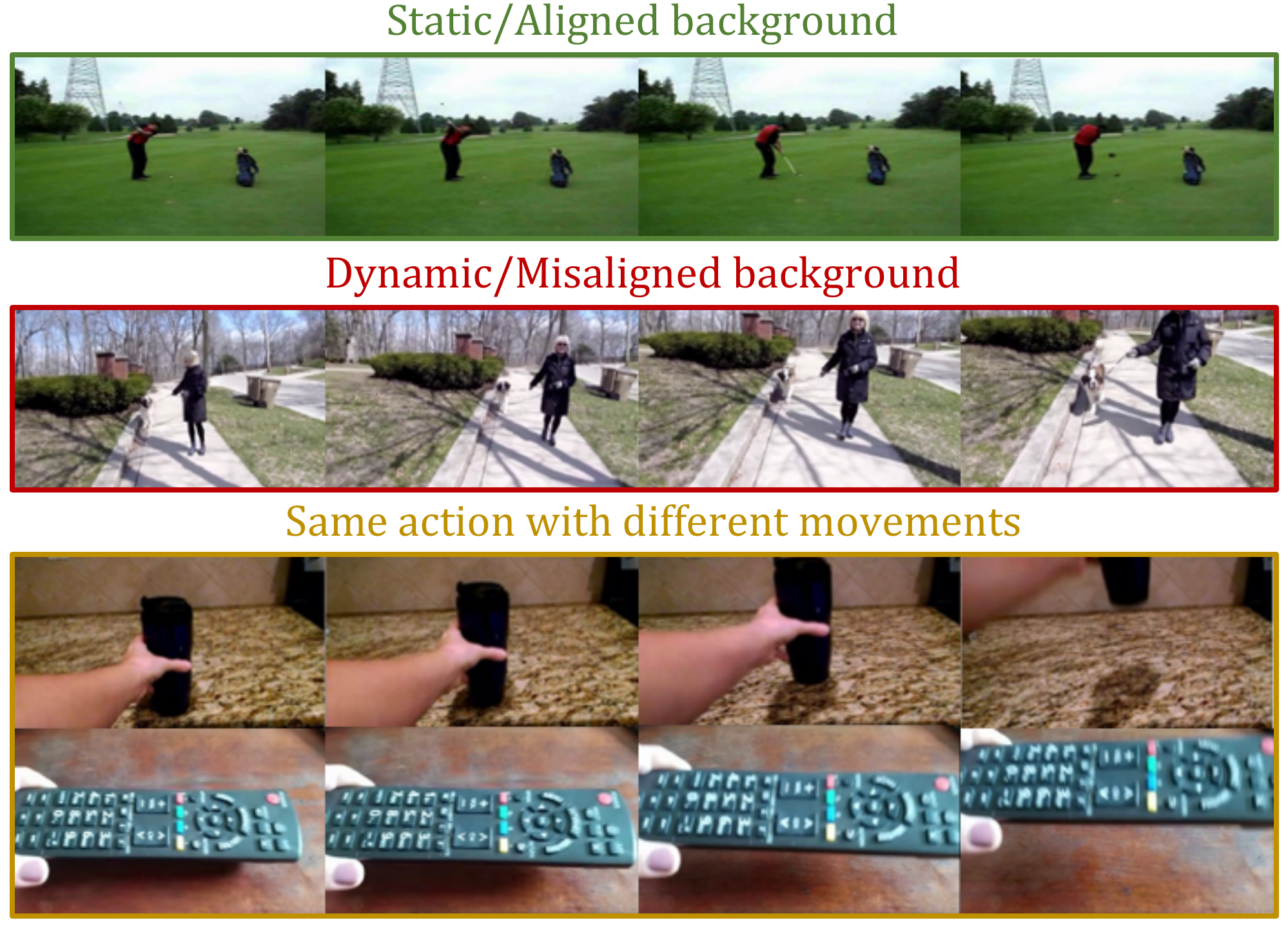}
\caption{\textbf{Illustration of our motivation.} First row is an example of ``golf " action with static/aligned background. Second row is the ``walk dog" action with dynamic/misaligned background owing to the camera movement. Third row shows the action (i.e., taking something from somewhere) with large movement, while the last row indicates the same action with slight movement.}
\label{fig:dataset}
\vspace{-0.3cm}
\end{figure}

Recent years have witnessed significant progress achieved in action recognition~\cite{twostream,stm,csn,su2018cascaded,su2020bsn++,su2020collaborative}. Deep learning based paradigm typically contains two main categories, i.e. two-stream networks~\cite{tsn,twostream,su2020transferable,sudhakaran2020gate} and 3D networks~\cite{c3d,csn,21D,x3d,fnet}. The former type of method captures the appearance and motion information from RGB images and stacked optical flow respectively. However, the extraction of optical flow is expensive in both time and space, which limits its real application. 3D convolutional networks exploit 3D convolutions to learn spatiotemporal features directly from raw videos. With the help of large-scale video datasets, superior performance can be obtained. However, tremendous parameters of stacked 3D convolutions inevitably increase the computing cost. 

Currently, many researchers explore to efficiently encode spatiotemporal features and motion encoding into a unified framework through decoupling 3D convolution into (2+1)D convolutions (e.g., $1\times3\times3$ and $3\times1\times1$) or performing feature-level difference between adjacent frames to capture motion dynamics with a simple convolution. Nevertheless, as shown in Fig.~\ref{fig:dataset}, videos taken by handheld devices can result in dynamic/misaligned background owing to the camera movements. Besides, the same action performed in different situations will lead to different movements (e.g., spatial shift and action speed). Hence, there still exists three main drawbacks: (1) direct feature-level difference without saliency alignment will inevitably introduce action-irrelated noises because of camera movements; (2) only local regional motion modeling is incapable of handling actions with various spatial shifts; (3) simply stacking local 1D convolutions without cross-perception integration is inferior to achieve multi-perception temporal modeling for various speed actions.

%current motion modeling methods lack the capability to model various movements because of limited receptive field.

%stacking local 1D convolutions is inferior to model long-range temporal relationships in the shallow layers of deep networks.
 
% Introduction of proposed method
To relieve the above issues, we propose a Temporal Saliency Integration (TSI) network to perform salient motion excitation and multiple perception temporal modeling for action recognition. Firstly, a Salient Motion Excitation (SME) module is proposed to capture comprehensive motion dynamics through spatial-level local-global motion modeling. Taking the camera movements into consideration, we perform the saliency alignment operation between adjacent frames to align the background in advance. Then the pyramidal motion modeling is conducted to handle various scales of action movements, thus obtain high-quality motion representations. Unlike \cite{stm}, which adds short-term motion encoding to the path of spatiotemporal features for the sake of complementary action recognition, we generate a channel-wise motion attention weight instead to enhance the motion-sensitive area of residual features as ~\cite{tea}. Next, in order to model multiple perception temporal relationships, a Cross-perception Temporal Integration (CTI) module is designed to implement multi-perception temporal modeling through adopting a series of depth-wise 1D convolutions upon the grouped channels respectively. Meanwhile, temporal interactions across multiple perceptions are integrated with attention mechanism, which is useful for ``local-to-global" and ``global-to-local" temporal information exchange. Finally, the proposed TSI block can be inserted into the off-the-shelf 2D CNNs to form a intuitive but effective network for comprehensive action recognition. Benefiting from the elaborate design, only limited extra computing cost is introduced (1.03$\times$ GFLOPs as many as 2D ResNet). Extensive experiments reveal that these two modules are complementary in long short-term temporal representation learning. To summarize, the main contributions of our work are three folds:

\begin{itemize}
\item The Salient Motion Excitation (SME) module is proposed to encode motion structure of various scales with fewer noises through saliency alignment and pyramidal motion modeling.
\item The Cross-perception Temporal Integration (CTI) module is designed to perform multi-perception temporal modeling with cross-perception integration. The two modules can be inserted into the standard ResNet block to collaboratively capture the long short-term temporal relationships.
\item Extensive experiments are conducted on five public benchmarks, which demonstrate that the proposed TSI network can outperform other state-of-the-art methods with superior performance and achieve competitive efficiency compared to 2D CNNs.
\end{itemize}

\section{Related Work}
%\textcolor{red}{simplify the description, cite more work like CT-Net}

\textbf{3D CNNs.} 3D convolution is naturally powerful for temporal modeling.
C3D~\cite{c3d} directly adopts the 3D convolutions to extract spatiotemporal features from raw videos. In order to make full use of the existing pre-trained 2D CNN on ImageNet\cite{imgnet} and thus expand the video modeling capabilities of the existing 2D CNN model\cite{dff,fernando2015modeling,smn}, I3D~\cite{i3d} proposes an inflate 2D convolution kernel as a 3D one to model the spatiotemporal features of video sequences. Since the spatial and temporal modeling process of 3D convolutions are highly coupled, it is not conducive for intuitive analysis of temporal modeling. 
Hence some researchers~\cite{cao2012scene,wu2018compressed,wu2019adaframe,x3d} propose to decouple the 3D convolution into (2+1)D convolutions~\cite{21D}. CSN~\cite{csn} is designed to substitute a normal 3D convolution with a depth-wise 3D convolution with another point-wise convolution to store the channel interactions. Recently, SlowFast \cite{slowfast} adopts two-stream 3D CNNs (e.g., slow and fast paths) to perform temporal fusion of different sampling speed.
However, these methods require large computation for training.

\begin{figure*}[t]
\centering
\includegraphics[width=1.95\columnwidth]{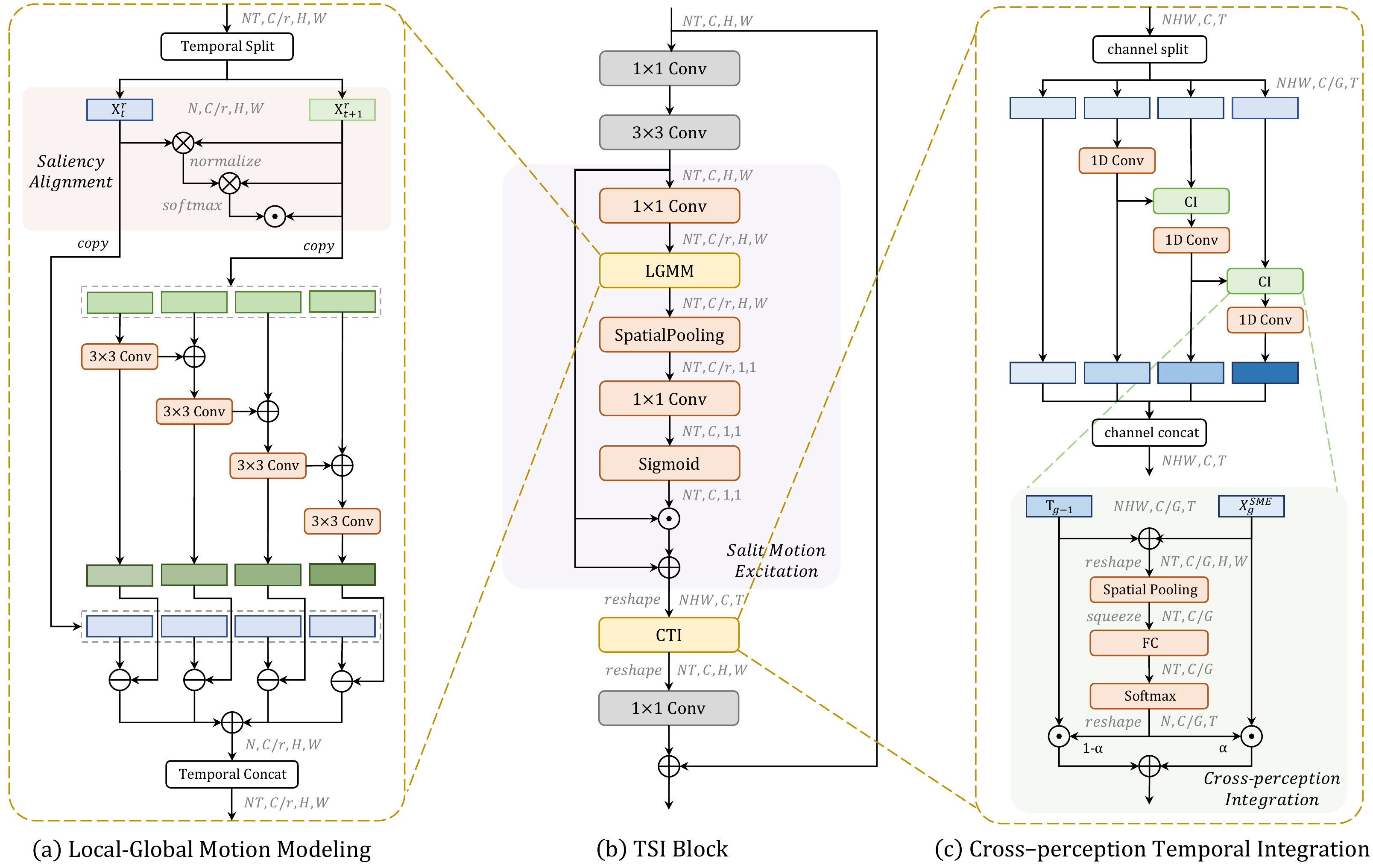}  
\caption{\textbf{Illustration of the proposed TSI block, which contains Salient Motion Excitation (SME) and Cross-perception Temporal Integration (CTI) modules.} In SME, saliency alignment between adjacent frames is conducted before performing pyramidal motion modeling, then the extracted motion features attend to the residual path after global average pooling, channel recovery and Sigmoid function. In CTI, multi-perception temporal modeling is performed in a hierarchical manner with cross-perception temporal integration.}
\label{fig:framework}
\vspace{-0.3cm}
\end{figure*}

\textbf{2D CNNs.} Recent works for video understanding mainly target on improvements of efficiency based on pre-trained 2D CNN. 
Two-stream network~\cite{two} adopts the spatial stream to extract appearance features and the temporal stream to capture motion information respectively. Then TSN~\cite{tsn} proposes a sparse sampling strategy and weighted average score fusion of these two streams. TRN~\cite{trn} exploits a set of fully connected layers to combine the relationships among frames with different intervals at the end. Recently, TSM~\cite{tsm} proposes a temporal shift operation to efficiently exchange channel information along the temporal dimension of features. STM~\cite{stm} adopts feature difference of neighboring frames to obtain the complementary short-term motion encoding of spatiotemporal features. TEINet~\cite{teinet} first performs feature-level spatial pooling and then adopts the adjacent feature difference to obtain channel-wise attention aiming to enhance the motion-sensitive features. TEA~\cite{tea} mainly employs inter-frame difference for short-term motion modeling and the Res2Net structure to model the long-term temporal relationships. Furthermore, CT-Net~\cite{tin} proposes to decompose different dimensions simultaneously after channel tensorization. 

However, without saliency alignment, action-irrelated noises will be introduced through direct feature-level temporal difference owing to the camera movement. Besides, only local regional motion modeling is incapable of handling different actions with various spatial shifts. Meanwhile, temporal interactions across different perceptions are also neglected in previous methods. In this paper, we propose Temporal Saliency Integration (TSI) network to handle these issues accordingly with the proposed Salient Motion Excitation (SME) and Cross-perception Temporal Integration (CTI) modules.

\section{Our Approach}
In this section, we will introduce the proposed Temporal Saliency Integration (TSI) network. First, we give an overview of the two main containing modules (i.e., Salient Motion Excitation and Cross-perception Temporal Integration). Then, we describe the technical details of these two modules respectively. Finally, we integrate them into a standard ResNet-50 backbone as TSI network.

\subsection{Overview}

As shown in Figure~\ref{fig:framework} (b), the proposed SME-CTI cube is the main component of TSI block which is designed for capturing short-term motion dynamics as well as long-term temporal inter-dependencies respectively. Briefly, the proposed SME-CTI cube has two main contributions. On one hand, Salient Motion Excitation (SME) module adopts pyramidal motion modeling to capture comprehensive motion dynamics of various actions. In addition, different from~\cite{tea,stm}, we introduce saliency alignment to eliminate the action-irrelated noises caused by misaligned background. On the other hand, Cross-perception Temporal Integration (CTI) module aims to capture multiple perception temporal relationships with cross-perception integration.
 
Specifically, the SME-CTI cube consists of SME and CTI. Given input feature $\mathbf{X} \in \mathbb{R}^{T \times C \times H\times W}$, where $T, C, H, W$ denote the number of frames, the number of filters, height and width of input features, respectively. $\mathbf{X}_{t}$\\ indicates the features of $t$-th snippet. We adopt SME to capture comprehensive motion dynamics between adjacent frames, thus enhancing the discriminability of motion-sensitive areas of input $\mathbf{X}$. Details will be described in Sec.~\ref{sec:SME}.

\iffalse
%which is formulated as follows:
\begin{equation}
    \mathbf{X}^{SME} = \text{SME}(\mathbf{X}),
\end{equation}
where SME represents the module used for extracting both local and global motion dynamics between adjacent frames, thus enhancing the discriminability of motion-sensitive areas of input feature $\mathbf{X}$. Details will be described in Sec.~\ref{sec:SME}.
\fi

Then we further adopt CTI for multiple perception temporal modeling. Concretely, a series of depth-wise 1D convolutions are performed upon several splitted channel groups respectively. Meanwhile, a cross-perception integration method is utilized to integrate temporal interaction between different temporal perceptions with channel-wise cross attention. Elaborate design contributes to the overall high efficiency without temporal diversity loss. Details will be described in Sec.~\ref{sec:CTI}.
\iffalse
\begin{equation}
\mathbf{X}^{CTI}=\text{CTI}(\mathbf{X}^{SME}),
\end{equation}
where CTI represents the Cross-perception Temporal Integration module, which performs upon the input features $\mathbf{X}^{SME}$. 
\fi

\subsection{Salient Motion Excitation}
\label{sec:SME}
As shown in Figure~\ref{fig:dataset}, different actions of inter / intra class can vary greatly, meanwhile, the background can also be misaligned owing to the camera movement, SME is thus proposed to encode the motion structure of various actions with less action-irrelated noises through saliency alignment and pyramidal motion modeling as shown in Figure~\ref{fig:framework} (a).

\begin{figure}[t]
\centering
\setlength{\abovecaptionskip}{0.1cm} %缩小caption和图像之间的距离
\includegraphics[width=1.0\columnwidth]{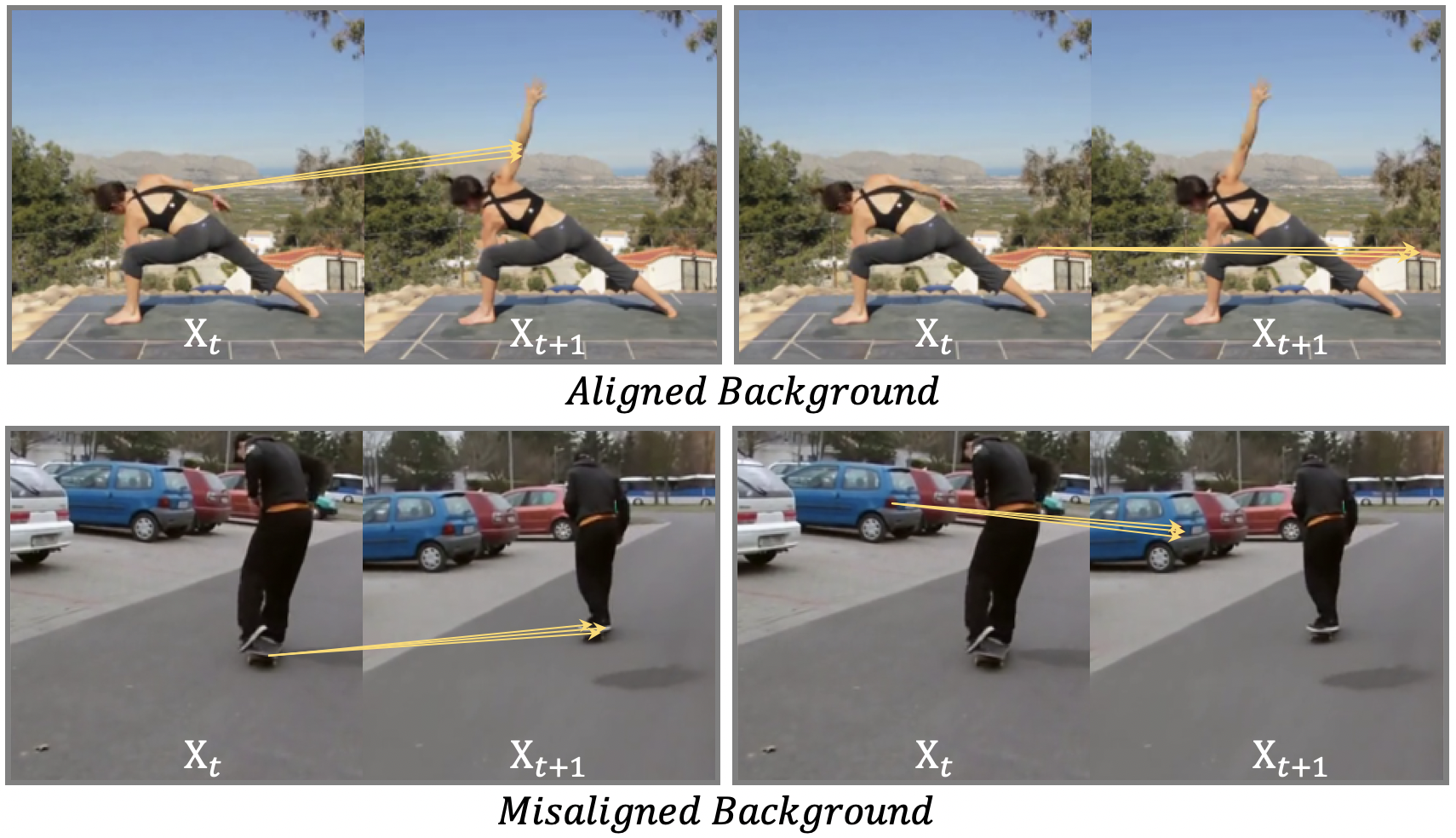}
\caption{\textbf{Visualization of saliency alignment.} 
The first row indicates the example with static/aligned background, and the second row indicates the example with dynamic/misaligned background. 
We select two points from foreground and background as anchors respectively
and show the top 3 points with the largest attention value in the next frame.
It shows that our method can dynamically align the corresponding context according to the similarity.}
\label{fig:visualization2}
\vspace{-0.3cm}
\end{figure}

\noindent
\textbf{Saliency Alignment.} Saliency alignment is designed for eliminating the action-irrelated noises introduced by misaligned background, thus contributing to the subsequent motion modeling process. Specifically, scaled dot-product is adopted for salient attention $\mathrm{S}(\mathbf{X}_{t}, \mathbf{X}_{t+1})$ calculation between adjacent frames, which is then adopted to highlight action-related regions of current frames $\mathbf{X}_{t+1}$ through element-wise multiplication. As shown in the upper part of Figure~\ref{fig:framework} (a), the features of $t$-th frame $\textbf{X}^{r}_{t}\in R^{N\times C/r\times H\times W}$ is adopted as query, and the features of next frame $\textbf{X}^{r}_{t+1}$ is adopted as key and value separately. The salient attention and alignment operators can be formulated as follow respectively:
\begin{small}
\begin{equation}
    \mathrm{S}(\cdot) =  Softmax(\mathbf{X}_{t}^{r} \otimes (\mathbf{X}_{t+1}^{r})^{T}/\sqrt{d}) \otimes \mathbf{X}_{t+1}^{r},
\end{equation}
\begin{equation}
    \mathbf{X}_{t+1}^{r(sa)} = \mathbf{X}_{t+1}^{r} \odot \mathrm{S}(\mathbf{X}_{t}^{r}, \mathbf{X}_{t+1}^{r}).
\end{equation}
\end{small}

$\otimes$ is the inner product, $d\!=\!C/r$ and  $\sqrt{d}$ is normalizing the inner product matrix row-wise, and $\odot$ is the hadamard product. $\mathbf{X}_{t+1}^{r(sa)}$ represents the output features after saliency alignment. The scaled and normalization steps are omitted in the Figure~\ref{fig:framework} (a) for clearness.
Figure~\ref{fig:visualization2} show representative examples of saliency alignment.

% (i.e., the similar parts inside the actors or the \textit{static} features of current frame $\mathbf{X}_{t+1}$) 

\noindent
\textbf{Discussion.}
Current methods \cite{stm,tea} adopt direct feature-level differences between neighboring frames to capture the short-term motion dynamics, which makes sense assuming the background is aligned. However, as shown in Figure~\ref{fig:dataset}, without considering the camera movements, the misaligned action-irrelated regions will be unexpectedly retained and the extracted motion representations are thus noisy and inferior. Saliency alignment is designed for eliminating the action-irrelated noises introduced by misaligned background, hence only the action-related regions are kept. Then the retained regions proceed to the following pyramidal motion modeling process to generate channel-wise attention weights aiming to highlight distinct salient regions. 
%are transformed into spatial-wise attention weights to highlight the distinct parts while suppress the irrelevant noises for later high-quality feature difference and motion extraction.
%  the salient parts inside the actors or aligned features of background
Figure~\ref{fig:visualization2} shows that
% where misaligned regions (i.e moving background) are effectively suppressed as shown in the first row, while the aligned regions (i.e., salient actors and static background) between neighboring frames are highlighted as shown in the second row.
even when the background is moving,
our saliency alignment module is able to relate similar semantic contexts,
e.g.,
the skateboard and the taillight.
%(i.e., moving edges of actors or \textit{dynamic} background) 

\iffalse
\begin{figure}[t]
\centering
\includegraphics[width=1.0\columnwidth]{Top.png}
\caption{Feature visualization of SME module. First row illustrates the input frames. The input and the output features of SME of Conv2$\_$1 block are presented in the second row and the third row respectively. Motion-sensitive areas (yellow dotted box) can be effectively highlighted while background noises (red dotted box) can also be suppressed.}
\label{fig:visualization}
\end{figure}
\fi

\noindent
\textbf{Pyramidal Motion Modeling.}
After saliency alignment, we continue to carry out pyramidal motion modeling for capturing spatial-level local-global motion dynamics. Considering the diversity of various actions, instead of using a simple convolution as in \cite{stm,tea,teinet}, we adopt a series of depth-wise 2D convolutions to model the spatial difference upon four copies of $\mathbf{X}^{r(sa)}$ successively, thus the equivalent receptive field of spatial dimension is enlarged in a hierarchical manner accordingly. Then the motion representations can be obtained through neighboring feature differences:
 
\begin{small}
\begin{equation}
    \mathbf{D}_{k}=\left\{
\begin{aligned}
 \mathrm{Conv2D}_{j}(\mathbf{X}_{t+1}^{r(sa)})&, &j=1,k=1 \\
 \mathrm{Conv2D}_{j}(\mathbf{D}_{k-1}+\mathbf{X}_{t+1}^{r(sa)})&, &2\leq j,  2 \leq k
\end{aligned}
\right.
\end{equation}
\end{small}
\begin{small}
\begin{equation}
    \mathbf{M}_{t, t+1} = \mathrm{Sum}(\mathbf{D}_{k}-\mathbf{X}_{t}^{r}), \quad k=1, ..., K.
\end{equation}
\end{small}

$\mathbf{M}_{t,t+1}$ is the feature difference between adjacent frames. And $K$ is set to 4 empirically. In order to reduce the computing cost, we adopt two 1$\times$1 2D convolutions for channel dimension reduction and recovery before and after Local-Global Motion Modeling respectively as shown in Figure~\ref{fig:framework} (b), where the ratio $r$ is set to 16. $\mathrm{Conv2D}_{j}$ indicates the $j$-th depth-wise 2D convolution used to transform the input features $\mathbf{X}^{r(sa)}$. Finally, the obtained local-global motion dynamics are further conducted spatial-level Global Average Pooling and Sigmoid activation function to obtain channel-wise attention weights $\mathrm{Att(\cdot)}$. The motion-sensitive areas of input features $\mathbf{X}$ are highlighted with element-wise multiplication and the background information is retained with an identity shortcut, where $\odot$ is the Hadamard product:
\begin{small}
\begin{equation}
    \mathrm{Att}(\cdot) =  Sigmoid(\mathrm{GAP}(\mathrm{LGMM}(\mathbf{X}_{t},\mathbf{X}_{t+1}))),
\end{equation}
\end{small}
\begin{small}
\begin{equation}
    \mathbf{X}^{\mathrm{SME}} =\mathbf{X} + \mathbf{X} \odot \mathrm{Att(\mathbf{X}_{t},\mathbf{X}_{t+1})}, t = 0, 1, ..., T-1.
\end{equation}
\end{small}
%\end{align}

\noindent
\textbf{Discussion.}
The proposed SME is different from the ME module proposed in TEA~\cite{tea} in two main aspects. On one hand, the saliency alignment is considered to eliminate the effect of misaligned regions caused by camera movements for effective motion modeling. Besides, pyramidal motion modeling is presented to enlarge the receptive field of spatial dimension for comprehensive motion extraction, which is essential but neglected in previous methods.

%Different from CMM module proposed in \cite{stm}, our SME module transforms the extracted motion features into attention weights to highlight the motion-sensitive regions, rather than adopts the motion encodings as another independent features. Besides, our 

\subsection{Cross-perception Temporal Integration}
\label{sec:CTI}
The structure details of CTI are shown in Figure~\ref{fig:framework} (c). The core idea of CTI is to model long short-term temporal information by multi-perception temporal modeling. Specifically, temporal receptive fields are equivalently enlarged as well as the temporal interactions are effectively exchanged through cross-perception integration.
 
\noindent
\textbf{Multi-perception Temporal Modeling.} Given the input feature $\mathbf{X}^{\mathrm{SME}} \in \mathbb{R}^{T\times C\times H\times W}$, we first separate the feature channels into $G\!\!=\!\!4$ groups, where $\mathbf{X}^{\mathrm{SME}}_{g} \in \mathbb{R}^{T\times C/G\times H\times W}$ indicates the $g$-th group features. Then we use a series of depth-wise 1D convolutions of kernel size 3 to extract temporal information of each feature group respectively. The temporal receptive field is equivalently enlarged and the multiple perception temporal information can be captured through the Cross-perception Integration (CI) method, which can be formulated in the following.
\begin{small}
\begin{equation}
  \mathbf{T}_{g}=\left\{
\begin{aligned}
  \mathbf{X}^{\mathrm{SME}}_{g}  & , & g=1 \\
 \mathrm{Conv1D}(\mathbf{X}^{\mathrm{SME}}_{g}) & , & g=2 \\
 \mathrm{Conv1D}_{j}(\mathrm{CI}( \mathbf{T}_{g-1},\mathbf{X}^{\mathrm{SME}}_{g} ))& , &3\leq j, g\leq 4
\end{aligned}
\right.
\end{equation}
\end{small}

$\mathbf{T}_{g}$ is the output temporal features of $g$-th group.

\noindent
\textbf{Cross-perception Integration.} In order to perform selective interaction fusion of multiple temporal perceptions, we first conduct spatial global average pooling on the summed features of adjacent groups. Then we adopt a fully connected layer followed by a Softmax activation function to obtain temporal-channel attention weights. Finally, the obtained weights are used to perform a weighted summation of temporal features with two different perceptions:
\begin{equation}
    \alpha = Softmax(\mathrm{FC}(\mathrm{GAP}(\mathbf{T}_{g-1}+ \mathbf{X}^{\mathrm{SME}}_{g} ))),
\end{equation}
\begin{equation}
    \mathrm{CI}(\mathbf{T}_{g-1}, \mathbf{X}^{\mathrm{SME}}_{g}) =\alpha \odot \mathbf{X}^{\mathrm{SME}}_{g} +(1-\alpha) \odot \mathbf{T}_{g-1}.
\end{equation}%

\noindent
\textbf{Discussion.} Different from the conventional temporal modeling methods \cite{stm,teinet}, which captures the long-term temporal information with stacking local 1D convolutions, CTI achieves this goal in a hierarchical manner. Meanwhile, unlike Res2Net structure adopted in \cite{tea}, which transfers the spatiotemporal information from local to global through single-pass element-wise addition, our CTI is designed to integrate ``local-to-global" and ``global-to-local" temporal information between different perceptions with the cross-attention mechanism. In this way, global temporal perception can be also exerted on the local details for better temporal modeling.

%In order to model long short-term temporal relationships within a single block, several 1D temporal convolutions with a kernel size of 3 are adopted for temporal modeling upon different channel groups respectively.  Previous method~\cite{tea} enlarges the temporal scale/receptive field of last channel groups through cascaded element-wise addition of temporal features, neglecting to integrate long-term temporal information into the short-term one. Differently, we adopt a cross attention method to perform cross-perception information exchange, thus the long-term temporal information can also be integrated into the short-term features gradually. In this manner, we can achieve the ``long-to-short" and ``short-to-long" temporal modeling intuitively.

\iffalse
\begin{figure}[t]
\centering
%\setlength{\abovecaptionskip}{-0.cm} %缩小caption和图像之间的距离
\includegraphics[width=1\columnwidth]{overview.png} % Reduce the figure size so that it is slightly narrower than the column. Don't use precise values for figure width.This setup will avoid overfull boxes.
\caption{The overall architecture of TSI network. The sparse sampling strategy~\cite{tsn} is adopted to sample $T$ frames from the input video. We adopt 2D ResNet-50 as the backbone, where the SME and CTI modules are inserted into each ResNet block to form the TSI block. The simple temporal average pooling is applied to conduct action predictions in the last score fusion stage.}
\label{fig:overview}
\end{figure}
\fi

\subsection{TSI Network}
To encode both spatiotemporal information and motion dynamics simultaneously, we combine the proposed SME and CTI together to form a TSI block, which can be easily inserted into the off-the-shelf 2D CNNs. In TSI block, the first 1$\times$1 2D convolution is adopted for channel reduction, then a 3$\times$3 2D convolution is utilized to model the spatial appearance, where the output features are further passed through SME and CTI sequentially to extract long short-term relationships. Finally, in order to restore the original channel dimensions, we add another 1$\times$1 2D convolution followed by a parameter-free identity shortcut from input to output. In order to give full play to the advantages of TSI block, from the perspective of easy optimization and flexibility, we use 2D ResNet-50~\cite{resnet} as our backbone unless specified. And we replace all standard ResNet blocks with our proposed TSI block.
%  as shown in Fig.~\ref{fig:overview}

%=============================================================Experiments==================================

\section{Experiments}
In this section, we first introduce the datasets and the implementation details of our proposed TSI. Then we perform extensive experiments to demonstrate that the proposed TSI  outperforms all the state-of-the-art methods on both temporal-related datasets and scene-related datasets. Meanwhile, we also conduct abundant ablation studies with Something-Something V1 to analyze the effectiveness of our method. Finally, we give runtime analyses to show the efficiency of TSI compared with state-of-the-art methods. 
\subsection{Datasets}
We verify the effectiveness of our proposed TSI on two kinds of datasets, which focus on temporal and context scenes respectively. First, \textbf{Something-Something V1 \& V2}~\cite{something} concentrate on temporal information, where V1 includes about 110$K$ videos while V2 includes 220$K$ video clips for 174 fine-grained classes. Second, for Kinetics-400, UCF-101 and HMDB-51, these datasets focus on the context scenes more. \textbf{Kinetics-400}~\cite{ki} has a total of 400 action classes, where the training set is about 240$K$ and the validation set is about 20$K$. The \textbf{UCF-101}~\cite{ucf} contains 101 categories with around 13$K$ videos, while \textbf{HMDB51}~\cite{hmdb} has about 7$K$ videos spanning over 51 categories.

\subsection{Experimental Setup}
\begin{table}[t]
\setlength\tabcolsep{3pt}   %% 这个就是重点！！
\setlength{\abovecaptionskip}{0.2cm} %缩小caption和图像之间的距离
% \vspace{-0.2cm}
\resizebox{\columnwidth}{!}{
    \centering
    \begin{tabular}{l|l|l|cc|cc}
    \Xhline{1.3pt}
    % \hline
    % \hline
    \multirow{2}*{Method} &\multirow{2}*{\#Frames}&\multirow{2}*{GFLOPs}& \multicolumn{2}{c|}{SSV1} &\multicolumn{2}{c}{SSV2} \\   
  %\cline{4-7}
    ~ & ~ & ~ & Top-1 & Top-5 & Top-1 & Top-5   \\
    \hline
    I3D\cite{i3d} & 32$\times$3$\times$2  & 153$\times$6 & 41.6 & 72.2 & N/A & N/A    \\
    I3D+GCN \cite{gcn} & 32$\times$3$\times$2   & 168$\times$6   & 43.4 & 75.1   & N/A & N/A \\
    ECO\cite{eco} & 8$\times$1$\times$1 & 32 & 39.6  & N/A & N/A & N/A        \\
    S3D-G\cite{s3d} & 64$\times$1$\times$1 & 71 & 48.2 & 78.7 & N/A & N/A   \\
    ir-CSN\cite{csn}  & 32$\times$1$\times$10 & 967$\times$10 & 49.3 & N/A & N/A & N/A  \\ 
    % \hline
    % \hline
    \Xhline{1.0pt}
    TSN\cite{tsn} & 8$\times$1$\times$1 & 33 & 19.7 & 46.6 & 27.8 & 57.6 \\
    TSM\cite{tsm} & 8$\times$1$\times$1 & 33 & 45.6 & 74.2 & 58.8 & 85.4 \\
    % TAM\cite{more_is_less} & 8$\times$1$\times$2 & 23.8 & 46.4 & 76.6  & 59.1 & 86.0 \\
    TEINet\cite{teinet} & 8$\times$1$\times$1 & 33 & 47.4 & N/A & 61.3 & N/A \\
    TANet\cite{tnnet} & 8$\times$1$\times$1 & 33 & 47.3 & 75.8 & 60.5 & 86.2 \\
    STM\cite{stm} & 8$\times$3$\times$10 & 33$\times$30 & 49.2 & 79.3 & 62.3 & 88.8 \\
    GSM\cite{sudhakaran2020gate} & 8$\times$1$\times$1 & 26.9 & 49.0 & 77.0 & N/A & N/A \\
    SmallBig\cite{smallbig} & 8$\times$2$\times$2 & 52$\times$6 & 48.3 & 78.1 & 61.6 & 87.7 \\
    TEA\cite{tea} & 8$\times$3$\times$10 & 35$\times$30 & 51.7 & 80.5 & N/A & N/A \\
    \hline
    Our TSI & 8$\times$1$\times$1 & 34 & 50.2 & 80.1 & 62.2 & 88.5 \\
    Our TSI & 8$\times$3$\times$10 & 34$\times$30 & \textbf{53.0} & \textbf{82.1} & \textbf{64.9} & \textbf{89.8} \\
    % \hline	    
    % \hline
    \Xhline{1.0pt}
    TSM\cite{tsm} & 16$\times$1$\times$1 & 65 & 47.3 & 77.1 & 61.2 & 86.9 \\
    % TAM\cite{more_is_less} & 16$\times$1$\times$2 & 47.7 & 48.4 & 78.8  & 61.7 & 88.1 \\
    TEINet\cite{teinet} & 16$\times$1$\times$1 & 66 & 49.9 & N/A & 62.1 & N/A \\
    TANet\cite{tnnet} & 16$\times$1$\times$1 & 66 & 47.6 & 77.7 & 62.5 & 87.6 \\
    STM\cite{stm} & 16$\times$3$\times$10 & 66$\times$30 & 50.7 & 80.4 & 64.2 & 89.8 \\
    GSM\cite{sudhakaran2020gate} & 16$\times$1$\times$2 & $53.7\times2$ & 51.7 & 79.6 & N/A & N/A \\
    SmallBig\cite{smallbig} & 16$\times$2$\times$2 & 105$\times$6 & 50.0 & 79.8 & 63.8 & 88.9 \\
    TEA\cite{tea} & 16$\times$3$\times$10  & 70$\times$30 & 52.3 & 81.9 & 65.1 & 89.9 \\
    \hline
    Our TSI & 16$\times$1$\times$1 & 68 & 53.3 & 82.5 & 63.2 & 89.1 \\
    Our TSI & 16$\times$3$\times$10 & 68$\times$30 & \textbf{54.3} & \textbf{83.2} & \textbf{66.1} & \textbf{90.9} \\
    % \hline
    % \hline
    \Xhline{1.3pt}
    \end{tabular}
}
% \vspace{-0.2cm}
\caption{\textbf{Comparison with the state-of-the-arts on Something-Something datasets.} 
N/A denotes the numbers are not available.}
\vspace{-0.3cm}
\label{table_comparison_sthv1}
\end{table}

\noindent
\textbf{Training.} We train our TSI with the same strategy as~\cite{tsn}. Given an input video, we first divide it into $T$ segments. Then we randomly sample one frame from each segment to obtain the input sequence with $T$ frames. The short side is fixed to 256. Meanwhile, corner-cropping and scale-jittering are applied for data augmentation. Finally, the input size of the network is $N\times T \times3 \times 224 \times 224$, where $N$ is the batch size and $T$ is the sampled frames per video. In our experiments, $T$ is set to 8 or 16 as default. For Kinetics and Something-Something datasets, the learning rate is set to 0.01 initially and then reduced by a factor of 10 at 30, 40, 45 epochs and stop at 50 epochs. We use the ImageNet~\cite{imgnet} pre-trained model as initialization on these datasets. As for UCF-101 and HMDB-51, we use Kinetics pre-trained model as initialization and set the learning rate to 0.001 for 25 epochs, which is decayed by a factor of 10 every 15 epochs. We use mini-batch SGD as the optimizer with a momentum of 0.9 and a weight decay of 5e-4. We train our TSI with 8 GTX 1080TI GPUs and each GPU processes a mini-batch of 8 videos clips (when T=8) or 4 video clips (when T=16).

\noindent
\textbf{Inference.} Following~\cite{tea}, we first re-scale the shorter spatial side to 256 and take three crops of $256\times256$ to cover the spatial dimensions and then resize them to $224\times 224$. We try two kinds of testing protocols: (1) 1 clip and center crop ($\times$1 view); (2) 10 clips and 3 crops ($\times$30 views).
%================================= Compare SOTA Method =====================
\subsection{Comparison with the State-of-the-Arts}
%We further demonstrate the advances of our proposed TSI in comparison with state-of-the-art methods for human action recognition.
For fair comparisons, all compared action recognition methods use only RGB modality as input with 8 or 16 frames. For UCF-101 and HMDB51, we use the testing protocols of 30 views, with 16 input frames as default. 
%The results on Something V1 \& V2, Kinetics-400, UCF-101 and HMDB-51 are shown in Tables~\ref{table_comparison_ucf_hmdb} respectively.
%======================== SV2 =================

\noindent
\textbf{Results on Something-Something.} We first verify the temporal modeling ability of TSI on Something-Something V1, and we can find that our TSI can achieve superior results to other methods as shown in Table~\ref{table_comparison_sthv1}. Specifically, with 16 frames as input, TSI is 2.0\% ahead of TEA \cite{tea} with fewer GFLOPs. Meanwhile, our method has obtained the advantage results than \cite{stm,teinet} under the same settings. Besides, with a single model, GSM\cite{sudhakaran2020gate} achieves 51.68\% Top-1 accuracy with 16×2 frames as input, while our proposed TSI method can obtain obviously higher accuracy (54.3\% vs. 51.68\%) with both fewer input frames (16×1) and fewer GFLOPs (68 vs. 107.4). For Something-Something V2, our TSI can still outperform the TEINet~\cite{teinet} by 1.1\% Top-1 accuracy with 16 input frames. Extensive experiments reveal that the temporal modeling ability of our TSI is both effective and efficient.

%========================= K400 =================
\begin{table}
\setlength\tabcolsep{3pt}   %% 这个就是重点！！
\setlength{\abovecaptionskip}{0.2cm} %缩小caption和图像之间的距离
% \vspace{-0.2cm}
\resizebox{\columnwidth}{!}{
    \centering
    \begin{tabular}{l|l|l|l|cc}
    \Xhline{1.3pt}
    Method & Backbone & \#Frames & GFLOPs & Top-1 & Top-5 \\
    \hline
    I3D\cite{i3d} & IncepV1 & 64$\times$N/A & 108$\times$N/A & 72.1 & 90.3 \\
    I3D-NL\cite{nl} & 3D R101 & 32$\times$3$\times$10  & 359$\times$30 & 77.7 & 93.3 \\
    ECO$_{en}$\cite{eco}  & BNIn+R18 & 92$\times$1$\times$1 & 267 & 70.7 & 89.4  \\
    S3D-G\cite{s3d} & IncepV1 & 64$\times$3$\times$10 & 71$\times$30 & 74.7 & 93.4 \\
    SlowFast\cite{slowfast} & 3D R50 & (4+32)$\times$3$\times$10 & 36.1$\times$30 & 75.6 & 92.1 \\
    SlowFast\cite{slowfast} & 3D R101 & (8+32)$\times$3$\times$10 & 106$\times$30 & \textbf{77.9} & \textbf{93.2} \\
    \Xhline{1.0pt}
    TSN\cite{tsn} & IncepV3 & 25$\times$1$\times$10 & 80$\times$10 & 72.5 & 90.2 \\
    R(2+1)D\cite{21D} & 2D R34 & 32$\times$1$\times$10 & 152$\times$10 & 72.0 & 90 \\
    TAM\cite{more_is_less} & 2D bLR50 & 48$\times$3$\times$3 & 93.4$\times$9 & 73.5 & 91.2 \\
    \Xhline{1.0pt}
    TSM\cite{tsm} & 2D R50 & 8$\times$3$\times$10 & 33$\times$30 & 74.1 &  N/A \\
    TEINet\cite{teinet} & 2D R50 & 8$\times$3$\times$10 & 33$\times$30 & 74.9 & 91.8 \\
    TEA\cite{tea} & 2D R50 & 8$\times$3$\times$10 & 35$\times$30 & 75.0 & 91.8 \\
    TDR\cite{tdr} & 2D R50 & 8$\times$3$\times$10 & 35$\times$30 & 75.7 & 92.2 \\
    \hline
    Our TSI & 2D R50 & 8$\times$3$\times$10 & 34$\times$30 & \textbf{75.9}  & \textbf{92.6}   \\
    \Xhline{1.0pt}
    TSM\cite{tsm} & 2D R50 & 16$\times$3$\times$10 & 66$\times$30 & 74.7 &  91.4 \\
    STM\cite{stm} & 2D R50 & 16$\times$3$\times$10 & 67$\times$30  & 73.7 & 91.6 \\
    TEINet\cite{teinet} & 2D R50 & 16$\times$3$\times$10 & 66$\times$30 & 76.2 & 92.5 \\
    TEA\cite{tea} & 2D R50 & 16$\times$3$\times$10 & 70$\times$30 & 76.1 & 92.5 \\
    TDR\cite{tdr} & 2D R50 & 16$\times$3$\times$10 &665$\times$30 & 76.9 & 93.0 \\
    TANet\cite{tnnet} & 2D R50 & 16$\times$3$\times$4 & 86$\times$12 & 76.9 & 92.9 \\
    \hline
    Our TSI & 2D R50 & 16$\times$3$\times$10 & 68$\times$30 & \textbf{77.1} & \textbf{93.7} \\
    \Xhline{1.3pt}
    \end{tabular}
}
% \vspace{-0.2cm}
\caption{\textbf{Comparison with the state-of-the-arts on Kinetics-400.} 
N/A denotes the numbers are not available.}
\label{table_comparison_k400}
\vspace{-0.1cm}
\end{table}

%3D and 2D methods, we can obtain better recognition results with limited additional computing cost.
%our method has less computational cost and reduce $1/5$ of the computation.

\setlength\tabcolsep{6pt}   %% 这个就是重点！！
\begin{table}[t]
\centering
\setlength{\abovecaptionskip}{0.2cm} %缩小caption和图像之间的距离
% \vspace{-0.2cm}
\resizebox{\columnwidth}{!}{
    \begin{tabular}{l|l|cc}
    \Xhline{1.3pt}
    Method  & Backbone   & UCF-101  & HMDB-51\\
    \hline
    C3D \cite{c3d}  & 3D VGG-11  & 82.3  & 51.6\\
    ECO \cite{eco}  & Incep+3D R18   & 94.8  & 72.4\\
    I3D \cite{i3d}  & 3D Inception-v1 & 95.1  & 74.3\\
    \Xhline{1.0pt}
    TSN \cite{tsn}  & ResNet-50   & 86.2  & 54.7\\
    TSM \cite{tsm} & ResNet-50  & 94.5 & 70.7\\
    STM\cite{stm}  & ResNet-50   & 96.2  & 72.2\\
    TEINet\cite{teinet}  & ResNet-50  & 96.7  & 72.1\\
    TEA\cite{tea}  & Res2Net-50   & 96.9  & 73.3\\
    MVFNet\cite{wu2020mv}  & ResNet-50 & 96.6  & 75.7\\
    \hline
    \textbf{TSI}  & ResNet-50  & \textbf{97.2} & \textbf{76.9}  \\
    \Xhline{1.3pt}
    \end{tabular}
}
% \vspace{-0.2cm}
\caption{\textbf{Comparison with the state-of-the-arts on UCF-101 and HMDB-51.} We load pre-train weights on Kinetics400.}
\label{table_comparison_ucf_hmdb}
\vspace{-0.3cm}
\end{table}

\begin{table*}[t]
    \setlength\tabcolsep{4pt}
    \begin{minipage}[t]{0.3\linewidth}
        \centering
        \vspace{0pt}
        \resizebox{\textwidth}{!}{
        \begin{tabular}{l|cc}
            \Xhline{1.0pt}
            Method & Top-1 & Top-5 \\
            \hline
            R50 + TIM & 46.1 & 74.7  \\
            R50 + MTA & 47.5 & 76.4 \\
            R50 + CTI  & \textbf{48.4} & \textbf{77.7} \\ 
            \hline
            R50 + TIM + ME  & 48.4      & 77.5 \\
            R50 + TIM + SME & \textbf{49.2} & \textbf{79.1} \\
            \Xhline{1.0pt}
        \end{tabular}
        }
        \subcaption{\textbf{Comparison with other temporal modules}. Both CTI and SME outperform previous modules for temporal modeling.}
    \label{table_temporal_ablation}
    \end{minipage}
    \hspace{0.5mm}
    \
    \setlength\tabcolsep{3pt}
    \begin{minipage}[t]{0.365\linewidth}
        \centering
        \vspace{0pt}
        \resizebox{\textwidth}{!}{
        \begin{tabular}{l|cc}
            \Xhline{1.0pt}
            Method & Top-1 & Top-5 \\
            \hline
            R50 + TIM & 46.1   & 74.7   \\
            \hline
            R50 + TIM + SME (w/o SA) & 48.3   & 77.9   \\
            R50 + TIM + SME (w/o CA) & 47.5   & 77.0   \\
            R50 + TIM + SME  & 49.2   & 79.1   \\
            \hline
            R50 + CTI & 48.4   & 77.7   \\
            R50 + CTI + SME & \textbf{50.2} & \textbf{80.1}   \\
            \Xhline{1.0pt}
        \end{tabular}
        }
        \subcaption{\textbf{Effectiveness of SME and CTI modules.}
        Utilizing both modules achieves the best performance.}
        \label{table_tet_ablation}
    \end{minipage}
    \hspace{0.5mm}
    \ 
    \setlength\tabcolsep{4pt}
    \begin{minipage}[t]{0.293\linewidth}
        \centering
        \vspace{0pt}
        \resizebox{\textwidth}{!}{
        \begin{tabular}{l|c|cc}%{L{4.8cm}R{1.7cm}R{1.7cm}}
            \Xhline{1.0pt}
            Location & Number & Top-1 & Top-5\\
            \hline
            None & 0 & 19.7 & 46.6 \\
            \hline
            Stage2 & 1 & 42.7 & 72.7 \\
            Stage3 & 1 & 44.5 & 74.4 \\
            Stage4 & 1 & 45.4 & 75.3 \\
            Stage5 & 1 & 45.5 & 75.0 \\
            \hline
            Stage2-5 & 4 & 51.8 &  81.2\\
            \hline
            Stage2-5 & 16 & \textbf{53.3} & \textbf{82.5} \\
            \Xhline{1.0pt}
            \end{tabular}
        }
        \subcaption{\textbf{Location and number of TSI block.}}
        \vspace{1mm}
        \label{table_location}
    \end{minipage}
    \
    \setlength\tabcolsep{7pt}
    \begin{minipage}[t]{1.0\linewidth}
        \centering
        \resizebox{\textwidth}{!}{
        \begin{tabular}{c|c|c|c|cc}%{L{4.8cm}R{1.7cm}R{1.7cm}}
            \Xhline{1.0pt}
            Module Fusion & Saliency Alignment & Motion Modeling & Cross-perception Integration & Top-1 & Top-5\\
            \hline
            \textit{\textbf{cascade}} & \textit{\textbf{element-wise multiplication}} & \textit{\textbf{pyramidal}} & \textit{\textbf{cross attention}} & \textbf{53.3} & \textbf{82.5} \\
            \textit{\textbf{summation}} & \color{lightgray}{\textit{element-wise multiplication}} & \color{lightgray}{\textit{pyramidal}} & \color{lightgray}{\textit{cross attention}} & 51.2 & 80.3 \\
            \textit{\textbf{concatenation}} & \color{lightgray}{\textit{element-wise multiplication}} & \color{lightgray}{\textit{pyramidal}} & \color{lightgray}{\textit{cross attention}} & 52.1 & 81.0 \\
            \color{lightgray}{\textit{cascade}} & \textit{\textbf{element-wise addition}} & \color{lightgray}{\textit{pyramidal}} & \color{lightgray}{\textit{cross attention}} & 52.7 & 82.0 \\
            \color{lightgray}{\textit{cascade}} & \color{lightgray}{\textit{element-wise multiplication}} & \textit{\textbf{simple}} & \color{lightgray}{\textit{cross attention}} & 52.2 & 81.3 \\
            \color{lightgray}{\textit{cascade}} & \color{lightgray}{\textit{element-wise multiplication}} & \color{lightgray}{\textit{pyramidal}} & \textit{\textbf{independent}} & 51.8 & 80.9 \\
            \color{lightgray}{\textit{cascade}} & \color{lightgray}{\textit{element-wise multiplication}} & \color{lightgray}{\textit{pyramidal}} & \textit{\textbf{element-wise addition}} & 52.5 & 81.8 \\
            \Xhline{1.0pt}
            \end{tabular}
        }
        \subcaption{\textbf{Detailed designs of TSI block.}
        All the designs bring significant performance gain.}
        \vspace{-0.2cm}
        \label{table_all_ablation}
    \end{minipage}
    \caption{\textbf{Ablation studies.} 
    Experiments are conducted on Something-Something V1 \cite{something}.
    We adopt ResNet50 + TIM~\cite{teinet} as our baseline by default.
    For Table (a) and (b), all the models are trained with 8 sampled frames,
    while 16 frames are sampled for other tables.
    }
    \vspace{-0.3cm}
    \label{ablation}
\end{table*}

\noindent
\textbf{Results on Kinetics-400.} As shown in Table~\ref{table_comparison_k400}, we also compare with other methods on Kinetics-400 to verify the model robustness in context scenario. Compared with all 2D methods, our TSI can achieve the state-of-the-art performance with competitive efficiency. As for 3D methods, TSI can also surpass most of the mainstream 3D methods with fewer input frames and less GFLOPs, i.e., I3D~\cite{i3d}, SlowFast~\cite{slowfast}, S3D-G~\cite{s3d} and ECO~\cite{eco}. Although the ensembled SlowFast~\cite{slowfast} model can achieve the a bit higher accuracy, our TSI has great efficiency with obvious fewer GFLOPs (68 vs. 106).

\noindent
\textbf{Results on UCF-101 and HMDB-51.} To verify the generalizability of our TSI network, we pre-train it on Kinetics-400 and then fine-tune it on UCF-101 and HMDB-51 respectively, the results are shown in Table~\ref{table_comparison_ucf_hmdb}. Compared with other methods, TSI can obtain consistently better performance on both datasets, with 97.2\% Top-1 accuracy on UCF-101 and 76.9\% Top-1 accuracy on HMDB-51.

\subsection{Ablation Studies}
In this section, we conduct exhaustive ablation experiments to evaluate the effectiveness of our TSI network on Something-Something V1 \cite{something}. Unless specified, the reported results are calculated using the testing protocol of a single view with 8 sampled frames.

\noindent
\textbf{Comparison with other temporal modules.} 
It is known that temporal modeling is vital for temporally related datasets.
% i.e., Something-Something V1\cite{something}.
In order to compare the temporal modeling ability with other state-of-the-arts, 
we adopt the ResNet50 equipped with TIM as our strong baseline as reported in Table~\ref{table_temporal_ablation}.
TIM indicates the temporal interaction module proposed in~\cite{teinet},
which adopts a depth-wise 1D temporal convolution with special initialization.
It clearly shows that via integrating multi-perception temporal information with cross attention,
our proposed CTI can also surpass the existing temporal modules (i.e., TIM \cite{teinet} and MTA \cite{tea}) by a great margin.
Meanwhile,
compared to ME proposed in~\cite{tea}, 
our proposed SME can obtain higher performance with better-extracted motion representations.
Therefore, 
we can conclude that the two proposed modules are both effective in long short-term temporal modeling.

\noindent
\textbf{Effectiveness of SME and CTI modules.} 
In Table~\ref{table_tet_ablation}, 
we study the effect of each individual module respectively. 
% In order to evaluate the temporal modeling ability of our TSI, we adopt the Res50 + TIM as our strong baseline on the temporally related dataset, where TIM indicates the temporal interaction module proposed in~\cite{teinet}, which adopts a depth-wise 1D temporal convolution. 
With SME,
the Top-1 accuracy can be boosted by 3.1\%, 
which reveals the effectiveness of local-global motion modeling. 
Besides, 
we also study the effect of Saliency Alignment (SA) and Channel-wise motion Attention (CA) operations adopted in SME module respectively. 
We can observe 0.9\% and 1.7\% drop of Top-1 accuracy respectively, 
demonstrating the necessity of each design choice. 
Moreover, 
our proposed CTI can achieve superior temporal modeling performance than TIM  (+2.3\%).
After combining these two modules, the recognition accuracy can be further improved by 1.8\%,
which confirms that these two modules are indispensable for comprehensive temporal representation learning.
 
\noindent
\textbf{Location and number of TSI block.}
Conventional ResNet50 architecture can be divided into 6 stages, where the last four stages (refer to Stage2-5) consist of several residual blocks respectively. In order to ensure the effectiveness of TSI block and the impact of the block number on the performance improvement, 
we replace the first residual block of each stage with our TSI block respectively, as shown in Table~\ref{table_location}.
Surprisingly,
we find that significant performance improvement already can be obtained even with only one inserted TSI block.
Besides, the inserted block in the latter stage can yield better performance than the early stage,
because the semantic features are more discriminative for temporal modeling and motion excitation in the deeper layers.
And inserting one TSI block into each stage (4 TSI blocks in total) is beneficial for further accuracy improvement.
Furthermore,
the best performance can be obtained through replacing all residual blocks with the proposed TSI block in Stage2-5 (16 TSI blocks in total).

\noindent
\textbf{Fusion of two proposed modules.} 
We further study the effect of different fusion types of two proposed modules as shown in row1-3 in Table~\ref{table_all_ablation}. 
Specifically, 
we try three fusion types respectively, 
namely summation, 
concatenation and cascade. 
The first two types indicate that the motion features and the multi-perception temporal features are extracted from the shared input in parallel and the outputs are element-wise summed or concatenated over channel dimension. 
While the cascade fusion type means that the two modules are conducted successively. 
We can observe that the last type can yield the best performance, 
which is mainly because that the importance of spatiotemporal features and motion encoding would not be equal in usual scenarios.

%Spatiotemporal and motion encodings are validated complementary in~\cite{stm}. And two different fusion types are evaluated (e.g. summation and concatenation) respectively, where the first type is chosen finally. Specifically, the spatiotemporal and motion features are extracted separately from the shared input by different modules, and the outputs are concatenated over the channel dimension or summed directly. Whatever fusion type is adopted, two branches are optimized independently. It is intuitive that the importance of spatiotemporal and motion features would not always be equal in usual scenarios. For example, as shown in Figure~\ref{fig:dataset}, the background information contributes a lot for determining the ``golf" action, while the temporal interactions are important for ``taking something from somewhere" action prediction. In this paper, we find that the cascade fusion can achieve the best results as shown in Table~\ref{table_fusion}, where the motion-highlighted features after SME module are fed to the following CTI module for temporal modeling. In this way, long short-term temporal modeling can be conducted consecutively and gradually.

\noindent
\textbf{Design choices of saliency alignment operation.} 
After salient attention calculation, 
we also explore two different saliency alignment operators, 
namely element-wise addition and multiplication.
As shown in the row1, 4 in Table~\ref{table_all_ablation},
the attended features conducted element-wise multiplication with the current features $\textbf{X}^{r}_{t+1}$ can obtain better performance than element-wise summation.
We argue that summation may introduce misalignment noise.

\noindent
\textbf{Study on pyramidal motion modeling.}
The key idea of our local-global motion modeling process is pyramidal motion modeling.
Compared with the previous methods~\cite{teinet,stm,tea},
which only adopt a simple convolution to perform motion difference modeling and feature transformation,
our TSI devises a pyramidal structure aiming to handle actions with various shapes and scales.
The spatial receptive field of the convolution kernel can be equivalently enlarged in a hierarchical fashion.
Results shown in row1, 5 in Table~\ref{table_all_ablation} confirm the significance of the hierarchical receptive field for our SME module.

\noindent
\textbf{Design choices of cross-perception integration.} 
Finally, 
we explore the effect of design choices of our cross-perception integration method as shown in row1, 7-8 in Table~\ref{table_all_ablation}. 
Without cross integration between each two neighboring channel groups (i.e., independent handling different groups),
the temporal modeling performance drops obviously owing to the lack of multiple perception information. 
The previous method~\cite{tea} adopts element-wise addition to transfer knowledge from local perception to global perception.
However, 
it is unable to exert global perception to the local one,
resulting in inferior performance.
These results verify the power of our CTI module.

\setlength\tabcolsep{5pt}   %% 这个就是重点！！
\begin{table}[tbp]
\setlength{\abovecaptionskip}{0.2cm} %缩小caption和图像之间的距离
% \vspace{-0.2cm}
\resizebox{\columnwidth}{!}{
    \centering
    \begin{tabular}{l|l|l|cc}%{L{2.2cm}C{1.5cm}C{1.5cm}C{1.5cm}R{1.5cm}}
    \Xhline{1.3pt}
    Method & \#Frames & GFLOPs & Throughput & Top-1 \\
    \hline
    I3D~\cite{i3d}  & 32$\times$3$\times$2    & 153$\times$6  & 6.1vid/s  & 41.6      \\
    ECO~\cite{eco}    & 16$\times$1$\times$1   & 64  & 45.6vid/s & 41.4      \\
    TSN~\cite{tsn}    & 8$\times$1$\times$1  & 33   & 81.5vid/s & 19.7      \\
    \hline
    \multirow{2}{*}{TSM~\cite{tsm}}     & 8$\times$1$\times$1     & 33   & 77.4vid/s & 45.6      \\
        & 16$\times$1$\times$1  & 65   & 39.5vid/s & 47.3      \\
    \hline
    \multirow{2}{*}{STM~\cite{stm}}    & 8$\times$1$\times$1    & 33   & 62.0vid/s & 47.5 \\
       & 16$\times$1$\times$1   & 66 & 32.0vid/s & 49.8     \\
     \hline
    \multirow{2}{*}{TEINet~\cite{teinet}}  & 8$\times$1$\times$1    & 33   & 46.9vid/s & 47.4    \\
     & 16$\times$1$\times$1   & 66   & 24.2vid/s & 49.9     \\
    \hline
    \multirow{2}{*}{TSN~\cite{tsn} + CTI}    & 8$\times$1$\times$1     & 33   & 61.6vid/s & 48.4      \\
      & 16$\times$1$\times$1    & 66   & 31.2vid/s & 51.2     \\
    \hline
    \multirow{2}{*}{\textbf{our TSI} }   & 8$\times$1$\times$1     & 34   & 44.0vid/s & 50.2     \\
       & 16$\times$1$\times$1    & 68   & 23.7vid/s & \textbf{53.3}     \\
    \Xhline{1.3pt}
    \end{tabular}
}
\caption{\textbf{Efficiency and accuracy comparison of TSI with other state-of-the-art methods on Something-Something V1.} ``vid/s" indicates number of processing videos per second. TSI beats all competing methods with 44 videos per second using 8 frames as input. Measured on a single NVIDIA GTX 1080TI GPU.}
\vspace{-0.3cm}
\label{table_qualitative}
\end{table}
 
\subsection{Runtime Analysis}
We continue to compare the model efficiency with several state-of-the-arts on the Something-Something V1 dataset. All evaluations are conducted on one GTX 1080TI GPU. Specifically, we adopt the number of videos processed per second as an efficiency metric, termed as throughput, as shown in Table~\ref{table_qualitative}. For fair comparisons, we evaluate the efficiency of our method by evenly sampling 8 or 16 frames from an input video and then only adopt the single view for testing. Concretely, compared with existing 3D methods~\cite{i3d}, our proposed TSI network can significantly surpass them in both GFLOPs and throughput. Compared with ECO~\cite{eco}, our TSI network can achieve similar throughput with both higher accuracy (+8\%) and fewer input frames. As for 2D CNNs, though TSI obtains half throughput than TSN~\cite{tsn}, it can obtain 30\% accuracy improvement with similar GFLOPs. It should be noted that our CTI module can achieve remarkable performance improvement with a small computing cost. To conclude, our TSI network can strike a certain trade-off between efficiency and recognition accuracy.

\section{Conclusion}
In this paper, we propose Temporal Saliency Integration (TSI) network for video action recognition, which mainly consists of the Salient Motion Excitation (SME) and the Cross-perception Temporal Integration (CTI) modules. SME is designed to highlight motion-sensitive areas with fewer noises caused by misaligned background, while CTI aims to integrate multiple perception temporal relationships with cross attention. Combining these two modules into a unified ResNet bottleneck, long short-term temporal relationships can be effectively captured. Extensive experiments are conducted on several public benchmarks, which demonstrate that our TSI network can achieve remarkable performance improvement with competitive efficiency compared to 2D CNNs.

\noindent\textbf{Limitations.} In this paper, saliency alignment operation adopted for eliminating the action-irrelated noises brought by misaligned background is intuitive and non-trivial in high-quality motion extraction. However, computing cost of the scale-dot product used for salient attention calculation is somewhat heavy, especially for the features in the shallow layers with high resolution, which hinders the convergent speed of the whole network to some extent. In the future work, alternative operators (e.g., multi-head self-attention) can be explored for efficient alignment.
 
{\small
\bibliographystyle{ieee_fullname}
\bibliography{egbib}
}

\end{document}